\documentclass[letterpaper, 10 pt, conference]{ieeeconf}  

\IEEEoverridecommandlockouts                              
\label{key}
\usepackage{graphics} 
\usepackage{epsfig} 
\usepackage{amsmath} 
\usepackage{amssymb}  
\usepackage{multirow}
\usepackage{array}
\usepackage{algorithmic}
\usepackage{caption}
\usepackage{colortbl}
\usepackage{multicol}
\usepackage{adjustbox}
\usepackage{tabularx}
\usepackage{booktabs}
\usepackage{url}
\usepackage{doi}

\usepackage{subfig}
\usepackage{floatrow}
\usepackage{multirow}

\newcolumntype{Y}{>{\raggedright\arraybackslash}X}
\newcolumntype{b}{Y}
\newcolumntype{z}{>{\hsize=.55\hsize}Y}
\newcolumntype{s}{>{\hsize=.15\hsize}Y}

\title{\LARGE \bf
Depth Estimation on Underwater Omni-directional Images 
\\ Using a Deep Neural Network  
}

\author{Haofei Kuang$^{1}$, Qingwen Xu$^{1}$, S\"oren Schwertfeger$^{1}$ 
\thanks{$^{1}$Authors are with School of Information Science Technology of ShanghaiTech University
        {\tt\small <kuanghf, xuqw, soerensch>@shanghaitech.edu.cn}}%
}

\begin{document}

\maketitle
\thispagestyle{empty}
\pagestyle{empty}

\begin{abstract}
	
	In this work, we exploit a depth estimation Fully Convolutional Residual Neural Network (FCRN) for in-air perspective images to estimate the depth of underwater perspective and omni-directional images. We train one conventional and one spherical FCRN for underwater perspective and omni-directional images, respectively. The spherical FCRN is derived from the perspective FCRN via a spherical longitude-latitude mapping. 
	For that, the omni-directional camera is modeled as a sphere, while images captured by it are displayed in the longitude-latitude form. 
	Due to the lack of underwater datasets, we synthesize images in both data-driven and theoretical ways, which are used in training and testing. 
	Finally, experiments are conducted on these synthetic images and results are displayed in both qualitative and quantitative way. The comparison between ground truth and the estimated depth map indicates the effectiveness of our method.  
	
\end{abstract}


\section{Introduction}
\label{sec: intro}
Due to the properties of underwater environments, underwater perception is quite different from air. Images captured in underwater case usually look bluish or greenish. Besides, the underwater images are more blurred than that in air captured by the same camera due to turbidity. These reasons increase the difficulty of depth estimation from images. Thus many researchers put effort on the underwater image processing. For example, using dark channel priors is proposed to restore underwater images in \cite{drews2016underwater,luczynski2017underwater}, inspired by He et al.'s work on removing haze in air \cite{he2011single}. Pfingsthorn et al. implemented underwater image stitching based on spectral methods \cite{pfingsthorn2010maximum}, which are more robust to turbidity than feature based methods.  Besides image enhancement, some work focuses on depth estimation. Peng et al. exploited the relationship between depth and blurriness of underwater images to estimate depth \cite{peng2015single}. In addition, deep learning was also applied to estimate the depth of underwater images, for example, Li et al. used a convolution neural network (CNN) to generate relative depth, which was then one of the inputs for a color correction network \cite{li2018watergan}. 

In addition to normal pin-hole cameras, omni-directional cameras are becoming popular due to their large field of view (FOV). They have been widely used on ground robots \cite{argyros2005robot,benosman2000panoramic,lemaire2007slam}. Some research groups also studied omni-directional cameras for underwater use since they provide more information than perspective ones on object detection, localization and mapping. Boult designed an omni-directional video equipment and put it on dolphins to capture data \cite{boult2000dove}. In \cite{bosch2015omnidirectional}, Bosch et al. improved on-land omni-directional cameras for underwater use and proposed the method for camera calibration. 

In this paper, we aim to estimate the depth of underwater omni-directional images. In contrast to on-land scenarios,underwater depth estimation is more challenging due to scattering and absorption effect \cite{drews2013transmission, peng2015single} as mentioned above. In the very beginning, Eigen et.al.~\cite{eigen2014depth} proposed a two-stack convolutional neural network to estimate depth from single images. Later, many researchers improved the performance of depth estimation based on deep learning \cite{kuznietsov2017semi,laina2016deeper,liu2015deep}. We try to apply deep learning to estimate depth of omni-directional underwater images in this work to solve the difficulty in underwater scenario. Since deep learning is a data-driven way, a large amount of data is necessary. However, underwater images are hard to collect, especially omni-directional ones. Thus, we generate synthetic datasets based on available in-air datasets to handle this issue. 

\begin{figure}[tb]
	\centering
	\includegraphics[width=0.8\linewidth]{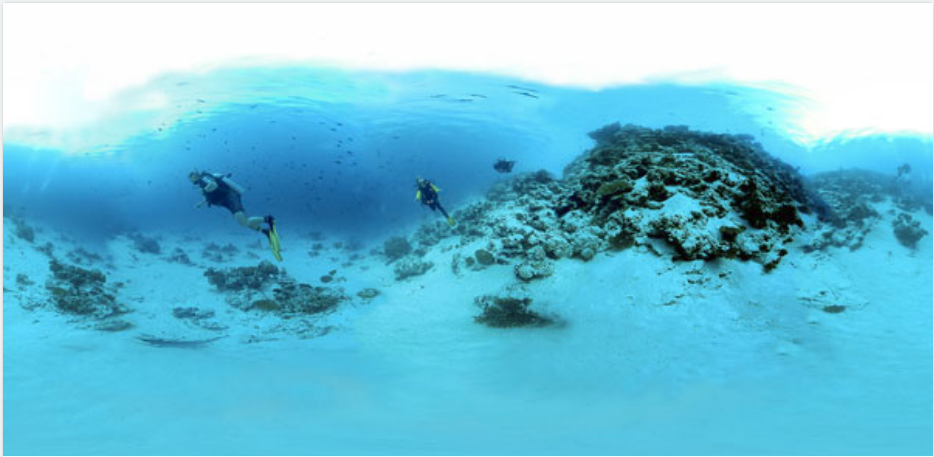}
	\caption{An underwater omni-directional image\protect\footnotemark.}
	\label{fig:UW_omni_sample}
\end{figure}

Another challenge is the serious distortion of omni-directional images  as shown in Figure~\ref{fig:UW_omni_sample}. For that, we learn from approaches that work for in-air images. Gehrig rectified the region of interest of omni-directional images to perspective images for the specific task of object detection\cite{gehrig2005large}. Omni-directional images are undistorted into longitude-latitude coordinates in \cite{li2008binocular}. Zhao et al. used a geodesic grid when extracting features on the omni-directional images \cite{zhao2015sphorb}. In \cite{adarve2017spherepix,ma20153d}, the omni-directional camera is modeled as a sphere and the omni-directional images are projected to the bounding cubic of the sphere, so that the image processing algorithms of perspective images, which are captured by pin-hole cameras, can be applied to the sub-images of each cubic side. In this work, we describe the omni-directional images in longitude-latitude coordinates and build a mapping between the longitude-latitude coordinate and the tangent space of the spherical camera model, then apply the mapping in deep neural networks as mentioned in \cite{coors2018spherenet,tateno2018distortion}.

\addtolength{\textfloatsep}{-.2in}

\footnotetext{\url{http://archive.doobybrain.com/2008/10/30/underwater-360-degree-vr-panoramas/}}

\addtolength{\textfloatsep}{.2in}

\begin{figure*}[tb]
	\centering
	\includegraphics[width=1.0\linewidth]{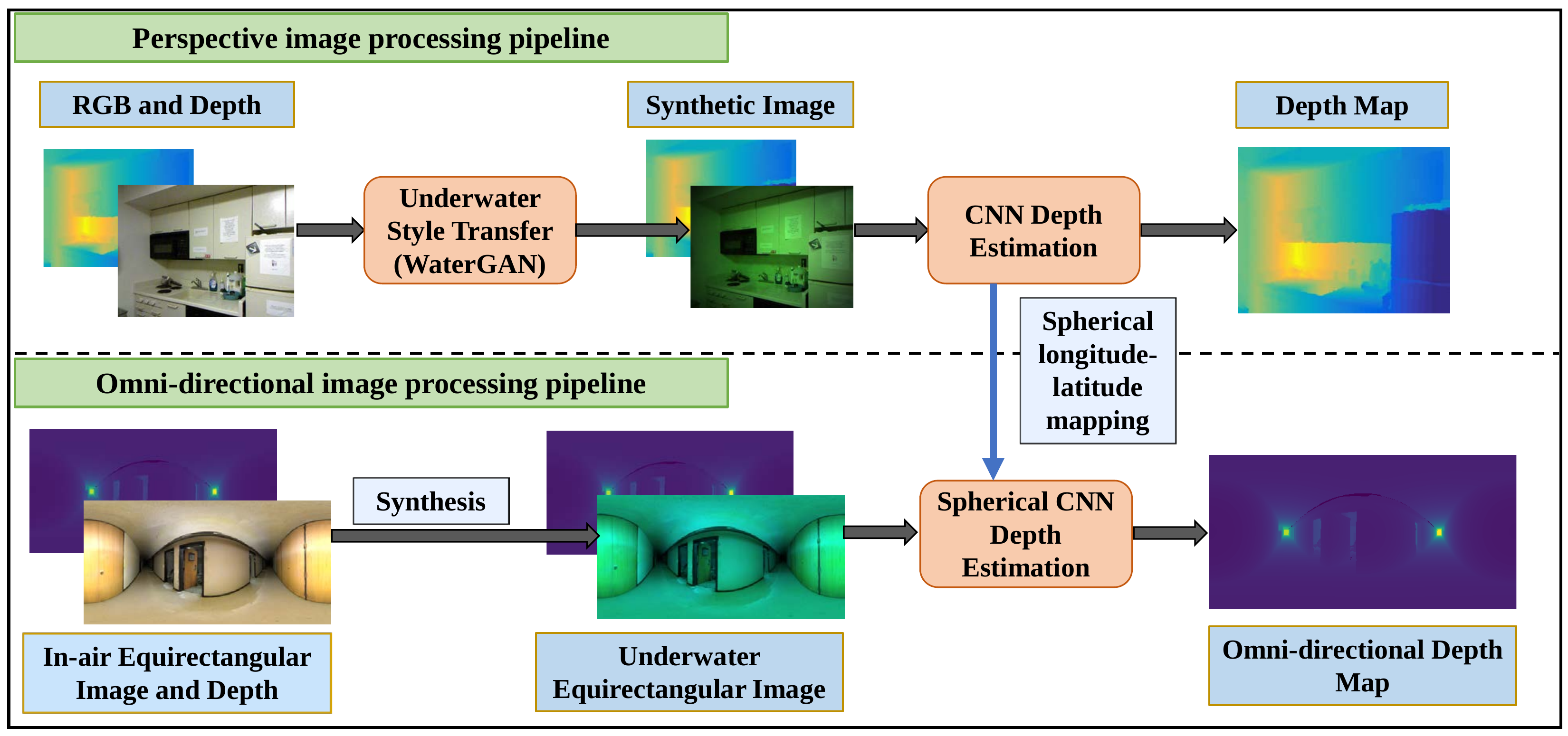}
	\caption{Perspective and omni-directional image processing training pipeline.}
	\label{fig:UW_system}

\end{figure*}

This paper is about transferring a neural network for in-air depth estimation to the underwater case. In Section~\ref{sec:construction} we introduce the pipeline of estimating depth on underwater images. Afterwards, all networks used in the work are explained in Section~\ref{sec:network}, where we also show approaches to the challenges of dataset generation and undistortion. In Section~\ref{sec:exp}, we show the experimental details and analyze the results. Finally, we conclude this work in Section~\ref{sec:conclusion}.


\section{System Overview}
\label{sec:construction}
 Applying deep learning, we conduct this work based on two known neural networks: WaterGAN \cite{li2018watergan} and FCRN \cite{laina2016deeper}. WaterGAN uses a generative adversarial network to transfer in-air perspective images to underwater tunes; FCRN has been proven to work well in depth estimation for both RGB and RGBD images. Furthermore, we use ideas presented in \cite{coors2018spherenet} to convert the networks to work with omni-directional images.

The pipeline of this work is shown in Figure~\ref{fig:UW_system}. The upper pipeline describes the training on perspective images. Firstly, RGB and depth images are transmitted into a style transfer network (WaterGAN). Then the synthetic underwater images and the corresponding depth images are the input of the conventional CNN (FCRN). After training, the network outputs the estimated depth map. The lower row shows the similar training process on omni-directional images, where we modify the conventional FCRN to spherical FCRN based on the similar idea introduced in \cite{coors2018spherenet,tateno2018distortion}. 
Since WaterGAN does not support style transfer on omni-directional images, we distort the in-air omni-directional images to underwater tunes by decreasing values in red channel and blurring images according to depth as mentioned in \cite{drews2016underwater}. Then the synthetic underwater omni-images are used as input to the spherical CNN network to estimate the omni-directional depth map, where the standard convolution and pooling is replaced with spherical ones.

\section{Neural Network Transfer}
\label{sec:network}
In this section, we introduce the two main neural networks used to solve the challenges: distortion removal, dataset generation and depth estimation. 

\begin{figure}[tb]
	\centering
	\includegraphics[width=0.8\linewidth]{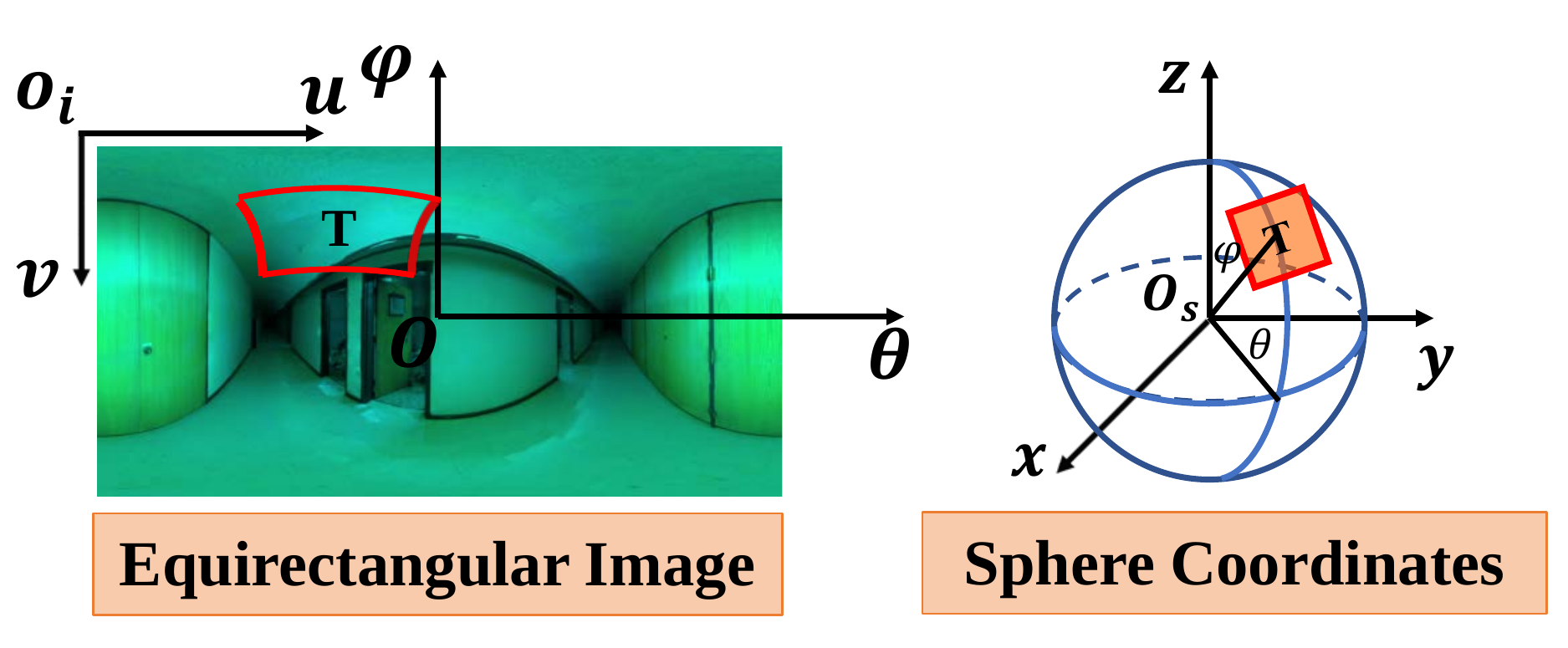}
	\caption{Mapping between longitude-latitude and spherical coordinates.}
	\label{fig:UW_Coor}
\end{figure}

\begin{figure*}[tb]
	\centering
	\sidesubfloat[]{
		\includegraphics[width=5cm]{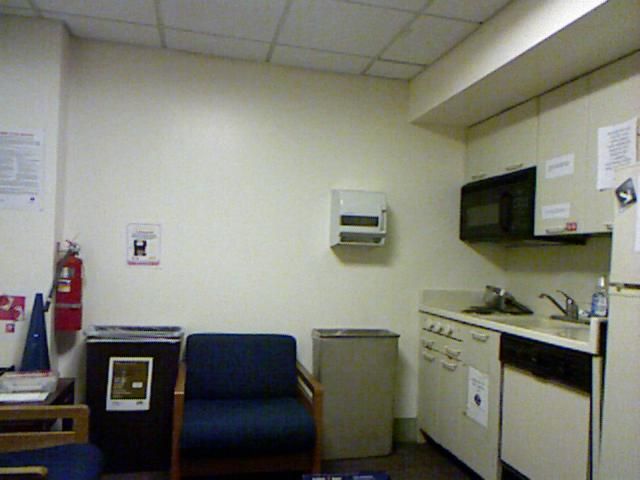} \quad
		\includegraphics[width=5cm]{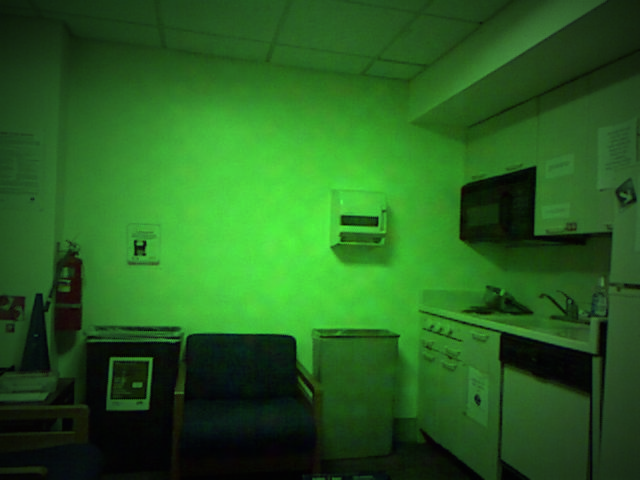} \quad
		\includegraphics[width=5cm]{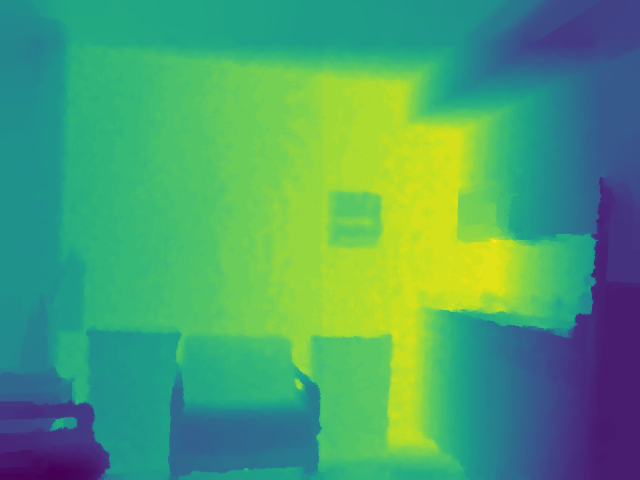}
		\label{fig:data:nyu}
	}
	\vspace{0.3cm}
	\sidesubfloat[]{
		\includegraphics[width=5cm]{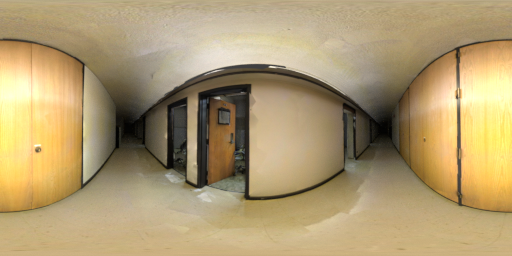} \quad
		\includegraphics[width=5cm]{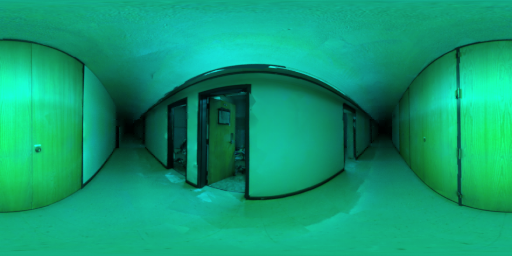} \quad
		\includegraphics[width=5cm]{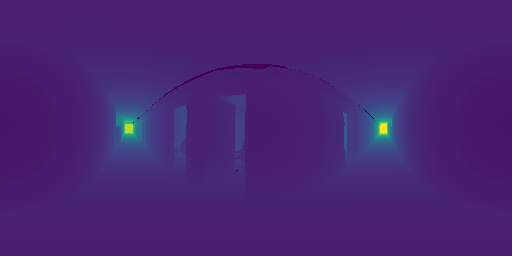}
		\label{fig:data:omni}
	}
	\caption{The examples from the NYU-v1 dataset and the 360D dataset. (a) is from the NYU-v1 dataset and (b) is from the 360D dataset. First column are RGB images from each dataset. Second column show the synthetic RGB images in underwater style. The last column shows the ground truth depth map.}
	\label{fig:data}
\end{figure*}

\subsection{Distortion Removal}
\label{ssec:distortion}
With the large FOV, omni-directional images suffer from serious distortions. We do not intend to remove all the distortion from the longitude-latitude rectangular image, but provide corrected pixel coordinates in the convolution neural network (CNN). In the convolution and pooling layer of CNN, a square kernel is used to slide over the image, which is not applicable on the equirectangular image, due to the distortion. Thus we exploit the sphere to model the omni-directional camera and build the mapping between the spherical surface $S(\phi, \theta)$ and equirectangular image $I(u,v)$. Then the image projected to the tangent space $T(x,y)$ of $S$ can be considered with no distortion \cite{adarve2017spherepix}. As shown in Figure~\ref{fig:UW_Coor}, the mapping between $I(u,v)$ and $S(\phi, \theta)$ can be described as 
\begin{subequations}
	\begin{equation}
		\phi = (\frac{H}{2} - v)\frac{\pi}{H}
	\end{equation}
	\begin{equation}
		\theta = (u - \frac{W}{2})\frac{2\pi}{W}
	\end{equation}
\end{subequations}
where $H$ and $W$ are the height and width of equirectangular image $I(u,v)$. Besides, the mapping between the tangent space $T(x,y)$ and spherical coordinates $S(\phi, \theta)$ can be described by gnomonic projection. Thus, when calculating the kernel coordinates during convolution and pooling, we only need to calculate the relative coordinates to center on $T(x,y)$. Extending from the $3\times 3$ kernel grid from \cite{coors2018spherenet}, the relative coordinates of $n\times n$ kernel can be described as $\left[\begin{array}{c}
	(\alpha x_{ij},\beta y_{ij})
	\end{array}\right]_{n\times n}$
when the kernel center coordinates is $(0, 0)$, where 
\begin{subequations}
	\begin{equation}
		x_{ij} = \tan( \left| i \right| \Delta_{\theta})
	\end{equation}
	\begin{equation}
		y_{ij} = \frac{\tan(\left| j \right| \Delta_{\phi})}{\cos( \left| i \right| \Delta_{\theta})}
	\end{equation}
\end{subequations}
, $\alpha$ and $\beta$ are the symbols consistent with the relative coordinates to center. 
Afterwards, we project these pixels $(x,y)$ on tangent plane $T(x,y)$ to $I(u,v)$ by inverse gnomonic projection.

\subsection{Synthetic Dataset}
\label{ssec:synthesis}
Even though there are some released underwater datasets, they cannot provide hundred thousands of images for the training of network. Motivated by \cite{li2018watergan}, we augment in-air images in underwater style to synthesize adequate underwater images for training. As mentioned in Sec~\ref{sec:construction}, perspective images are augmented in a data-driven way, i.e. use WaterGAN\footnote{\url{https://github.com/kskin/WaterGAN}} to transfer in-air images to underwater ones based on given underwater samples. And omni-directional images are distorted in color space and depth channel. The red light disappears firstly in the ocean due to its short wavelength so that the underwater image often looks blue or green as introduced in \cite{saeed2018Deep}. Besides, the underwater object becomes blurred when it gets far away from the camera owing to the attenuation of direct light and backscattering effect \cite{drews2016underwater,peng2015single}. Thus the in-air images can be converted to underwater style approximately by reducing red component of the images and blurring the images according to pixels' depth. This can be implemented in the following:  
\begin{subequations}
	\begin{equation}
		I_{w}(u, v, 'r') = \gamma * I_{a}(u, v, 'r')
	\end{equation}
	\begin{equation}
		I_{w}(u, v) = GaussBlur(I_{a}(u, v), f(d(u,v)));
	\end{equation}
\end{subequations}
, where $I_w$ and $I_a$ are underwater and in-air images, $0 < \gamma < 1$ is the attenuation factor of the red component, $I(u, v, 'r')$ represents the red channel of an image and $f(d(u,v))$ is the kernel size of Gaussian blur depend on pixel $(u,v)$ whose depth is $d(u,v)$. The synthetic perspective and omni-directional image samples are shown in Figure~\ref{fig:data:nyu} and \ref{fig:data:omni} respectively. Both underwater samples looks greenish than in-air ones.

\subsection{Depth Estimation}
\begin{figure*}[tb]
	\centering
	\includegraphics[width=1.0\linewidth]{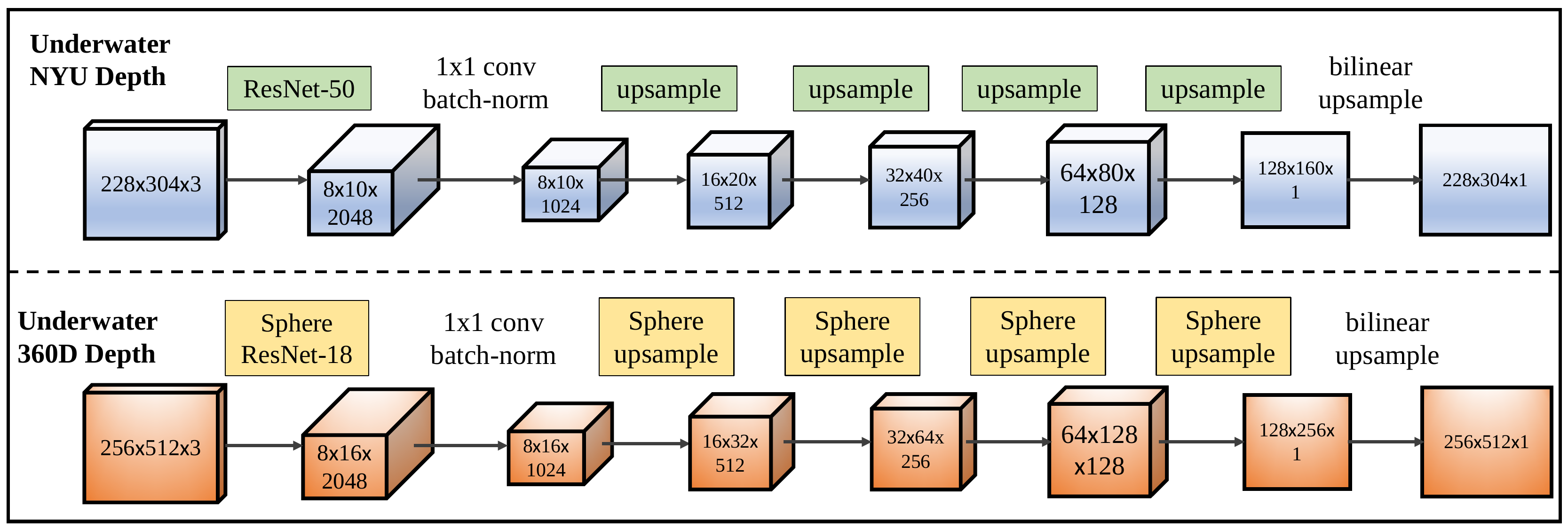}
	\caption{The architecture of depth estimation networks for perspective and omni-directional images. We train a ResNet-50 FCRN model for the underwater NYU dataset. We are then using a Sphere ResNet-18 architecture to train the model with underwater 360D datasets to reduce the memory and training time.}
	\label{fig:single_network}
\end{figure*}
The core idea of spherical CNN is introduced in both \cite{coors2018spherenet} and \cite{tateno2018distortion}. Since codes of both works have not been released, we modify a conventional CNN, FCRN\footnote{\url{https://github.com/iro-cp/FCRN-DepthPrediction}}, to realize our spherical CNN based on partial details from these works. 
The structure of standard and spherical CNN is shown in Figure~\ref{fig:single_network}. We use conventional FCRN to estimate depth for underwater perspective images and spherical FCRN for omni-directional ones. 

In the conventional FCRN, we use ResNet-50 \cite{he2016deep} as feature extraction layer (encoding layer) and up-project \cite{laina2016deeper} as up-sampling (decoding) layer. Besides, the unpooling layer of FCRN is replaced with deconvolution layer in order to simplify the calculation as reported in \cite{mal2018sparse}. In the spherical FCRN, the convolution and pooling layers are replaced with spherical convolution and pooling to build a SphereResNet model. 
In other words, we use the approach described in Sec~\ref{ssec:distortion} to calculate the corresponding pixels on omni-directional images with given pixels and square kernel. Then these corresponding pixels are used for spherical convolution and pooling calculation. Besides, ResNet-18 is the alternative to ResNet-50 in SphereResNet to reduce the consumption of memory and training time.

In addition, the mean absolute error $\mathcal{L}_1$ is used as default metric for its simplicity and performance, when operating optimization via Stochastic Gradient Descent (SGD) and back-propagation (BP). Moreover, we rescale the input images to meet the resolution consistency in \cite{tateno2018distortion} that the angle per pixel of both perspective and omni-directional images should be the same.

\section{Experiments and Results}
\label{sec:exp}
We perform the proposed method with synthetic datasets in the experiment. All networks: WaterGAN, FCRN and spherical FCRN are trained on a Titan V with 12G memory. 
The in-air dataset \textbf{NYUv1} \cite{silberman2011indoor} and real underwater images \textbf{MHL} \cite{li2018watergan} are used as the input of WaterGAN to synthesize underwater style \textbf{NYUv1} (\textbf{UW-NYU}).Thanks to the robustness of WaterGAN, we maintain the settings with learning rate $0.0002$, batch size 64 and training epoch 25. 

Both depth estimation networks are implemented with PyTorch\footnote{\url{https://pytorch.org/}}.
In the perspective images depth estimation, we train the FCRN model with the batch size of 16 on \textbf{UW-NYU}. The number of total training epochs is 30, the start learning rate is 0.01 and reduced 20\% every 5 epochs; weight decay is $ 1e^{-4} $ for regularization and momentum is 0.9. Afterwards, we build spherical FCRN based on the conventional one and the hyper-parameter are the same as FCRN. Finally, we use the synthetic underwater omni-directional images to train and validate the spherical FCRN.

In addition to show the advantage of our network, we also compare the performance of our network with Eigen et al.'s~\cite{eigen2014depth} on the perspective images.

\begin{figure}[tbp]
	\centering
	\sidesubfloat[]{
		\includegraphics[width=2.5cm]{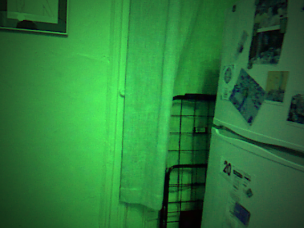}
		\hspace{0.01in}
		\includegraphics[width=2.5cm]{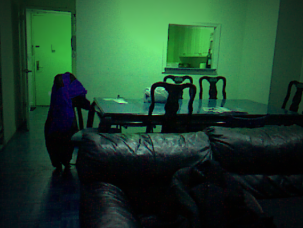}
		\hspace{0.01in}
		\includegraphics[width=2.5cm]{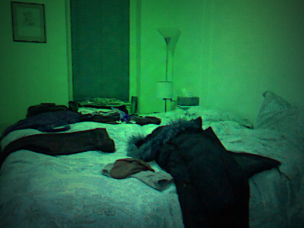}
		\label{fig:result:uw:rgb}
	}
	\vspace{.1in}
	\quad
	\sidesubfloat[]{
		\includegraphics[width=2.5cm]{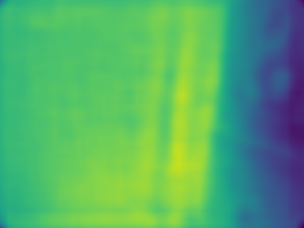}		\hspace{0.001cm}
		\includegraphics[width=2.5cm]{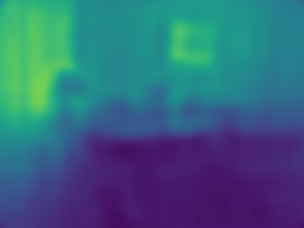}		\hspace{0.01in}
		\includegraphics[width=2.5cm]{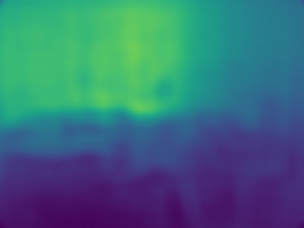}
		\label{fig:result:uw:pred_eigen}
	}
	\vspace{.1in}
	\quad
	\sidesubfloat[]{
		\includegraphics[width=2.5cm]{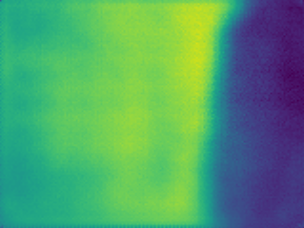}		\hspace{0.001cm}
		\includegraphics[width=2.5cm]{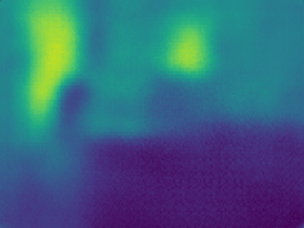}		\hspace{0.01in}
		\includegraphics[width=2.5cm]{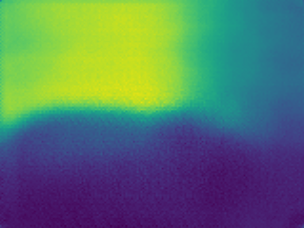}
		\label{fig:result:uw:pred}
	}
	\vspace{.1in}
	\quad
	\sidesubfloat[]{
		\includegraphics[width=2.5cm]{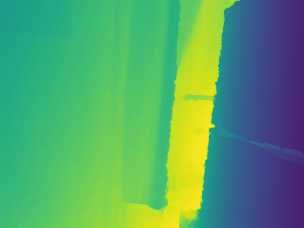}		\hspace{0.01in}
		\includegraphics[width=2.5cm]{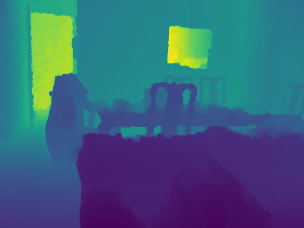}		\hspace{0.01in}
		\includegraphics[width=2.5cm]{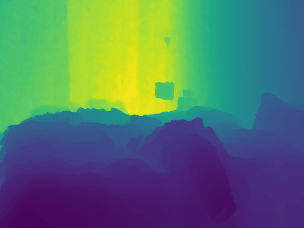}
		\label{fig:result:uw:gd}
	}
	\caption{The experimental results of underwater NYU dataset. (a) are RGB images from the testing set. (b) are the predicted depth of Eigen et al.'s~\cite{eigen2014depth}; (c) are the predicted depth maps of ours; (d) are the ground truth depth maps.}
	\label{fig:results:uw}
\end{figure}

\begin{figure*}[tbp]
	\centering
	\sidesubfloat[]{
		\includegraphics[width=5.3cm]{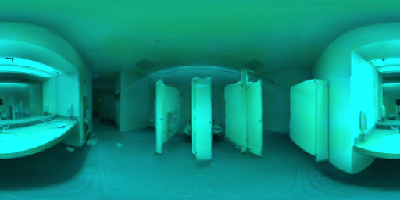}
		\hspace{0.01in}
		\includegraphics[width=5.3cm]{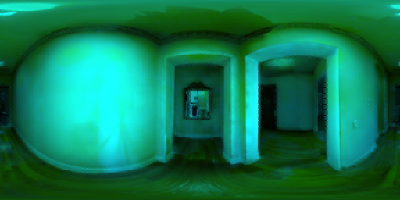}
		\hspace{0.01in}
		\includegraphics[width=5.3cm]{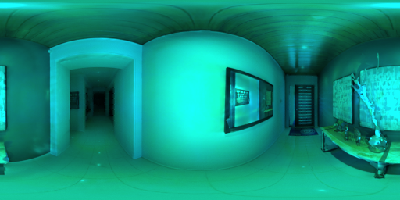}
		\label{fig:result:omni:rgb}
	}
	\vspace{.1in}
	\quad
	\sidesubfloat[]{
		\includegraphics[width=5.3cm]{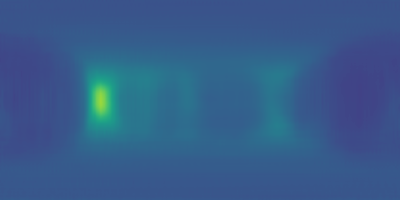}		\hspace{0.01in}
		\includegraphics[width=5.3cm]{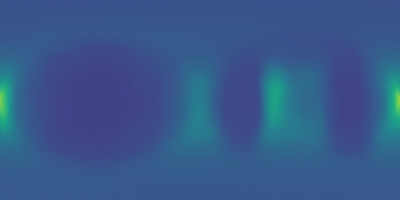}		\hspace{0.01in}
		\includegraphics[width=5.3cm]{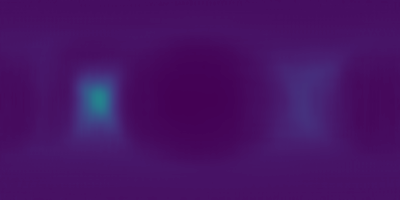}
		\label{fig:result:omni:pred}
	}
	\vspace{.1in}
	\quad
	\sidesubfloat[]{
		\includegraphics[width=5.3cm]{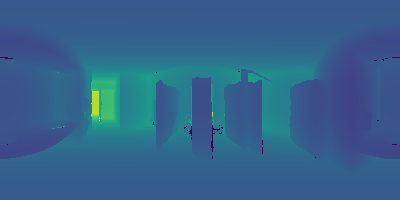}		\hspace{0.01in}
		\includegraphics[width=5.3cm]{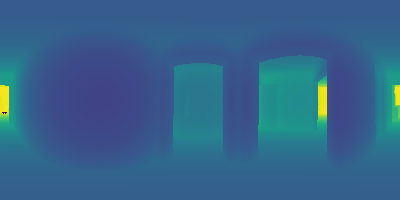}		\hspace{0.01in}
		\includegraphics[width=5.3cm]{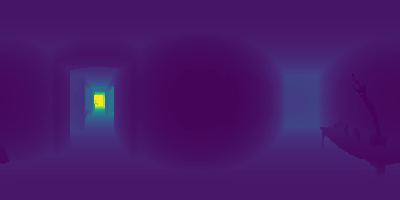}
		\label{fig:result:omni:gd}
	}
	\caption{The experimental results of underwater 360D dataset. (a) are RGB images from the testing set. (b) are the predicted depth maps; (c) are the ground truth depth maps.}
	\label{fig:results:omni}
\end{figure*}

\subsection{Data Augmentation}
To strengthen the robustness of the network, we conduct data augmentation on input images, including scale, rotation, flips, color jitter and normalization, similar to the procedure in \cite{mal2018sparse}. After augmentation, we crop the images from the center to keep the consistency of the input images. 



\subsection{Error Metrics}
We also take the same metrics in \cite{mal2018sparse} to evaluate our network. Here we address the metrics again:
\begin{itemize}
	\item RMSE: root mean squared error
	\item MAE: mean absolute error
	\item REL: mean absolute relative error
	\item $ \delta_1 = \frac{\#(\{\hat{y}_i:\max\{\frac{\hat{y}_i}{y_i}, \frac{y_i}{\hat{y}_i}\} < 1.25^{i}\})}{\#(\{y_i\})}$ ($ y_i $, $ \hat{y}_i $ are respectively the ground truth and the prediction) : percentage of predicted pixels whose depth error is smaller than a threshold. The higher $ \delta_1 $ is,  the better prediction is.
\end{itemize}

\begin{table*}[]
	\begin{tabular}{|c|c|c|c|c|c|c|}
		\hline
		Input & Model & RMSE & MAE & REL & $ \delta_1 $ & t\_gpu(s) \\ \hline
		\multirow{2}{*}{Perspective} & ResNet-50 & 0.162 & 0.117 & 0.098 & 0.914 & 0.0201 \\ \cline{2-7} 
		& Eigen et al. & 0.235 & 0.184 & 0.148 & 0.806 & 0.005 \\ \hline
		Omnidirectional & SphereResNet-18 & 0.604 & 0.362 & 0.172 & 0.711 & 0.0145 \\ \hline
	\end{tabular}
	\caption{The error metrics of each model. t\_gpu means the average operation time on each image on GPU.}
	\label{tab:results}
\end{table*}

\subsection{Results}
\subsubsection{Perspective Images} 
The experiment result of the depth prediction of \textbf{UW-NYU} dataset is shown in Figure~\ref{fig:results:uw}, and the precision evaluation is described in the second row of Table~\ref{tab:results}. The estimated depth map is close to ground truth except some details. 
Small RMSE, MAE and REL, together with $ \delta_1 $ beyond 0.9 of ours in Table~\ref{tab:results} indicates the result of the predict depth map achieve a high performance on the testing dataset, which is much better than Eigen et al.'s~\cite{eigen2014depth}. Besides, the average prediction time of each image is about $0.02s$. 

\subsubsection{Omnidirectional Images} 
Figure~\ref{fig:results:omni} shows the experimental results of the spherical FCRN, which still predicts depth correctly for most pixels. The quantitative results in the third row of Table~\ref{tab:results} also shows that the result is acceptable. However, RMSE, MAE and REL of spherical FCRN are higher than that of conventional FCRN and $\delta_1$ decreases a little in Table~\ref{tab:results}, which indicates that the performance is not as good as FCRN. The main reason is that we replace the ResNet-50 with ResNet-18 in spherical FCRN since the input omni-directional image is too big and our hardware cannot support training these images with ResNet-50.


\begin{figure}[tb]
	\centering
	\subfloat[RGB image]{
		\includegraphics[width=3cm]{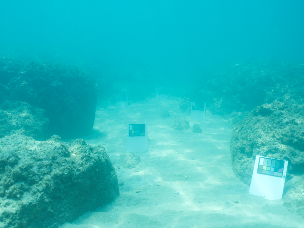}
		\label{fig:results:real_rgb}
	} \hspace{0.5cm}
	\subfloat[Depth map]{
		\includegraphics[width=3cm]{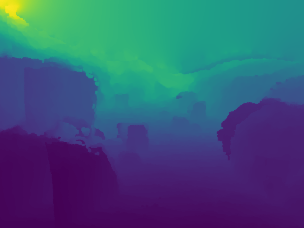}
		\label{fig:results:real_gd}
	}\\
	\subfloat[Predicted depth of Eigen et al.'s]{
		\includegraphics[width=3cm]{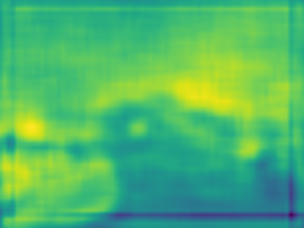}
		\label{fig:results:real_pred_eigen}
	}\hspace{0.5cm}
	\subfloat[Predicted depth of ours]{
		\includegraphics[width=3cm]{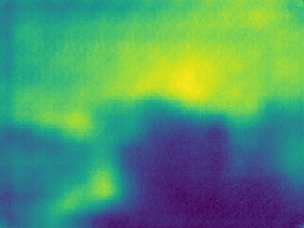}
		\label{fig:results:real_pred}
	}
	\caption{We test the trained FCRN model on a real underwater image which is proposed by Berman et al. \cite{berman2018underwater}. (a) and (b) are the input RGB image and depth map respectively. (c) is the predicted map of Eigen et al.'s~\cite{eigen2014depth}. (d) is the predicted map of ours.}
	\label{fig:results:real}
\end{figure}

\subsubsection{Testing on Real Data} 
In order to verify the feasibility of our method, we test our model with some real underwater data collected by Berman et al. \cite{berman2018underwater}. Figure~\ref{fig:results:real} shows a sample of the testing results, where the predicted depth is not correct in some pixels. However, our method still performs better than Eigen et al.'s. One possible reason could be the style of this image sets is different from our training dataset. The former looks blueish while the latter looks greenish. Another possible reason is the difference between camera models. The largest depth of training dataset is $10$ meters while the the longest distance of Figure~\ref{fig:results:real_gd} is more than 15 meters.  


\subsection{Discussion}
The above experimental results show that the conventional and spherical FCRN achieve good results on synthetic underwater perspective and omni-directional images, respectively. However, it does not show good performance when testing on a real underwater dataset. The main possible reason is that the style of the real underwater dataset and the synthetic dataset are too different. To overcome this challenge, we will enlarge the diversity of the training dataset in our future work, for example mixing images captured from different devices. Moreover, we will try to synthesize underwater images using accurate image formulation model \cite{Monika2019real}. 
On the other hand, we also plan to collect more real underwater omni-directional images to validate the spherical FCRN model. In addition, we will put more effort on transferring conventional FCRN to spherical FCRN without retraining, to save computation time and resources, which is reported feasible in \cite{tateno2018distortion}. 

\section{Conclusion}
\label{sec:conclusion}
We trained two depth estimation networks: FCRN for underwater perspective images and spherical FCRN for underwater omni-directional images, which are based on state-of-art depth estimation networks (FCRN) for in-air perspective images. Due to the lack of datasets, we synthesize underwater perspective images in a data-driven way and omni-directional images according to theoretical analysis. The test results on these synthetic images show that conventional and spherical FCRN can estimate depth map correctly on most pixels for synthetic underwater perspective and omni-directional images, respectively. In addition, we tested the trained FCRN model on real underwater images, which didn't give good results. To improve on that, we will increase the robustness of our network by collecting and more real underwater data and fine-tuning the estimation network based on the real data in our future work. 


\bibliographystyle{plain} 
\bibliography{references.bib} 
\end{document}